# Solar Flare Prediction Using Long Short-term Memory (LSTM) and Decomposition-LSTM with Sliding Window Pattern Recognition

Zeinab Hassani[1], Davud Mohammadpur[1], and Hossein Safari[2,3]
[1] Department of Computer Engineering, University of Zanjan, Zanjan, Iran; hassanizeinab@znu.ac.ir, dmp@znu.ac.ir
[2] Department of Physics, Faculty of Science, University of Zanjan, Zanjan, Iran; safari@znu.ac.ir
[3] Observatory, Faculty of Science, University of Zanjan, Zanjan, Iran

## Abstract

We investigate the use of long short-term memory (LSTM) and decomposition-LSTM (DLSTM) networks, combined with an ensemble algorithm, to predict solar flare occurrences using time series data from the GOES catalog. The data set spans from 2003 to 2023 and includes 151,071 flare events. Among approximately possible patterns, 7552 yearly pattern windows are identified, highlighting the challenge of long-term forecasting due to the Sun's complex, self-organized-criticality-driven behavior. A sliding window technique is employed to detect temporal quasi-patterns in both irregular and regularized flare time series. Regularization reduces complexity, enhances large flare activity, and captures active days more effectively. To address class imbalance, resampling methods are applied. LSTM and DLSTM models are trained on sequences of peak fluxes and waiting times from irregular time series, while LSTM and DLSTM, integrated with an ensemble approach, are applied to sliding windows of regularized time series with a 3 hr interval. Performance metrics, particularly the true skill statistic (0.74), recall (0.95), and the area under the curve (AUC = 0.87) in the receiver operating characteristic, indicate that DLSTM with an ensemble approach on regularized time series outperforms other models, offering more accurate large-flare forecasts with fewer false errors compared to models trained on irregular time series. The superior performance of DLSTM is attributed to its ability to decompose time series into trend and seasonal components, effectively isolating random noise. This study underscores the potential of advanced machine learning techniques for solar flare prediction and highlights the importance of incorporating various solar cycle phases and resampling strategies to enhance forecasting reliability.

*Unified Astronomy Thesaurus concepts:* Solar flares (1496); Solar x-ray flares (1816); Neural networks (1933); Time series analysis (1916)

## 1. Introduction

Solar flares are intense eruptions of electromagnetic radiation from the Sun's surface that travel the 150 million kilometers (1 au) to Earth in just 8 minutes. They emit radiation across the whole electromagnetic spectrum, with a concentration in ultraviolet and X-rays. Solar flares can disrupt radio communication systems, affect global navigation satellite systems, damage satellite equipment, interfere with bulk power grids on Earth, and pose risks to astronauts' health (D. M. Oliveira & C. M. Ngwira 2017; Z. M. K. Abda et al. 2020; V. Landa & Y. Reuveni 2022; S. Taran et al. 2023). These events can cause billions of dollars in damage and may require months for recovery. Consequently, accurate forecasting of solar flares is critical to mitigate their potentially destructive effects. In recent years, several approaches to forecasting solar flares have been explored, including time series analyses, automaton avalanche models, machine learning algorithms (both supervised and unsupervised), and parametric and nonparametric prediction methods (T. Colak & R. Qahwaji 2009; X. Huang et al. 2013; G. Barnes et al. 2016; J. Wang et al. 2020; A. Asensio Ramos et al. 2023; K. Dissauer et al. 2023; I. Kontogiannis 2023; K. D. Leka et al. 2023; S. Liu et al. 2023; M. K. Georgoulis et al. 2024).

Predicting solar flares based on sunspot classifications was one of the earliest approaches (P. S. McIntosh 1990). One efficient flare-forecasting method involved Bayesian analysis applied to flare statistics (M. S. Wheatland 2005). Various flare statistics approaches for observational data and simulations, such as the cellular automaton avalanche method, were developed to characterize the probability distribution of fluxes and waiting times within the flare system (A. Strugarek & P. Charbonneau 2014; N. Farhang et al. 2018, 2019; C. Thibeault et al. 2022). Additionally, solar flare forecasting has been explored using machine learning algorithms, including support vector machines (SVMs; Y. Yuan et al. 2010; M. G. Bobra & S. Couvidat 2015; L. E. Boucheron et al. 2015; N. Nishizuka et al. 2017; S. Asaly et al. 2021), random forests (C. Liu et al. 2017), light gradient boosting machines (P. A. Vysakh & P. Mayank 2023), logistic regression (H. Song et al. 2009), decision trees (D. Yu et al. 2009), artificial neural networks (O. W. Ahmed et al. 2013; S. Shin et al. 2016; K. Florios et al. 2018), long short-term memory (LSTM; Y. Chen et al. 2019), and deep neural networks (T. Nagem et al. 2018; N. Nishizuka et al. 2020, 2021; V. Landa & Y. Reuveni 2022; Y. Abduallah et al. 2023).

Y. Abduallah et al. (2023) proposed a framework called SolarFlareNet, based on transformers, to forecast γ-class solar flares in active regions 24 to 72 hr in advance. This framework includes three distinct transformers designed to predict large solar flares, such as M-, C-, and M5.0-class events. P. A. Vysakh & P. Mayank (2023) utilized the light gradient boosting machine, a machine learning model, to forecast class

C and M solar flares. Their study was based on a 9 yr data set from the Space-weather Helioseismic and Magnetic Imager Active Region Patches (SHARP). H. Baeke et al. (2023) employed magnetic parameters of active regions for solar flare prediction using an autoencoder. They also classified active regions using clustering techniques, including k-nearest neighbors, k-means, and Gaussian mixture models. Their findings showed that the resulting clusters included both low and high flare-producing regions, indicating that the SHARP data set lacks a clear boundary for effectively distinguishing flare-prone active regions. In another study, L. F. L. Grim & A. L. S. Gradvohl (2024) explored deep learning models based on active regions for forecasting ⩾M-class flares, using transformer-based models considering the sequence of magnetogram images. They also examined deep learning models for solar flare forecasting, utilizing transformer models that incorporate the sequence of magnetogram images from active regions. Various feature selection methods (e.g., Zernike moments and magnetic field parameters) were applied to extract essential features from solar extreme ultraviolet images and magnetograms (line of sight or SHARP), which were then used in different machine learning algorithms for solar flare prediction (A. Raboonik et al. 2016; N. Alipour et al. 2019; L. D. Krista & M. Chih 2021; S. Sinha et al. 2022; L. E. Boucheron et al. 2023; J. Wang et al. 2023; X. Huang et al. 2024). M. K. Georgoulis et al. (2024) presented an extensive review of solar flare forecasting, covering aspects such as the pre-flare/pre-eruption environment, methods for predicting solar flares and eruptions, a proposed framework for eruption forecasting, current challenges and objectives, and future needs and perspectives.

Flare-forecasting algorithms face challenges due to imbalanced data sets when extracting features from images, magnetograms, or time series data (e.g., GOES flux). In methods using image and magnetogram-based features, large flares and corresponding active regions are significantly less frequent than small flares and nonflaring active regions. Similarly, in time series data such as flux parameters, large flares (e.g., X class) are much less common than smaller ones (A and B classes), a phenomenon characteristic of a self-organized criticality system that follows a power-law distribution. The smaller population of significant flaring events, compared to small and midsized flares, creates a high imbalance in flare-forecasting algorithms, resulting in poorer performance (e.g., recall, precision, and F1-score) for the minority class in various machine learning models. When classifying imbalanced time series data, the model favors the majority class and fails to accurately identify the minority class (J. Liu et al. 2023). To address this issue, standard imbalance methods, such as resampling and cost-sensitive approaches, are recommended for solar flare forecasting (N. Moniz et al. 2017). One approach is resampling the data to balance the classes, while another method involves adjusting the statistical learning model to emphasize the minority class, for example, by applying class weights.

Extracting features from irregular time series, such as house prices and air quality, poses a significant challenge in neural network research (X. Liu et al. 2023). Among the crucial characteristics of these time series are the peak values and their corresponding waiting times, which play a vital role in understanding underlying patterns and trends. Sliding windows are a commonly used technique for analyzing time series data, particularly for detecting patterns and trends within localized segments. Typically, patterns exhibit varying lengths within a time series (M. Vafaeipour et al. 2014; J.-S. Chou & N.-T. Ngo 2016; G. Wang et al. 2018; Q. Dai et al. 2023). D. Janka et al. (2019) explored pattern detection and positional patterns identified by sliding windows in the context of steel production, with classification performed using LSTM and a temporal convolutional network (TCN). The spike sorting model, which employs a sliding window in conjunction with LSTM and 2D convolutional neural network (CNN) classification, segments data for spike detection and classification (M. Wang et al. 2023). More recently, sliding windows have been applied to flare forecasting (R. Tang et al. 2021; Y. Abduallah et al. 2023), with features such as magnetogram parameters considered based on their corresponding labels.

In this study, we explore the balancing of flare time series distribution using sliding window patterns. A smoothing algorithm is applied to the flare time series, and we analyze flare patterns recorded from 2003 to 2023 by the Geostationary Operational Environmental Satellites (GOES) using sliding window resampling. The sliding windows allow us to examine patterns in the imbalanced solar flare data, facilitating the estimation of different patterns for forecasting and analysis, even with a limited time series. For the irregular (original) flare time series, we apply LSTM and decomposition-LSTM (DLSTM) algorithms to sliding window quasi-patterns, which include both peak flux subwindows and waiting time subwindows, for flare forecasting. The LSTM model captures long-term and short-term dependencies in the flare time series, while the DLSTM model identifies local trends (seasonal variations) within the series. This forecasting method is employed for a binary classification problem, distinguishing between small flares with flux less than $10^{-5}\,\mathrm{W\,m^{-2}}$ (A, B, C classes) and large flares with flux greater than $10^{-5}\,\mathrm{W\,m^{-2}}$ (M and X classes).

For the regularized flare time series, we implement LSTM and DLSTM with ensemble algorithms using sliding windows that include only the peak flux subwindow for flare prediction. To mitigate class imbalance between small and large flares, regularizing the time series with a specific time interval is a key strategy to improve the prediction performance of large (minority class) flares.

We assess the effectiveness of six solar flare prediction models: (1) LSTM applied to irregular time series, (2) DLSTM on irregular time series, (3) LSTM on regularized time series, (4) DLSTM on regularized time series, (5) LSTM on regularized time series with ensemble learning, and (6) DLSTM on regularized time series with ensemble learning. To mitigate the class imbalance between small and large flares, we apply a resampling strategy to the minority class across all models. The models are evaluated based on performance metrics such as accuracy, precision, recall, true skill statistics (TSS), and Appleman skill score (ApSS). Additionally, we compute the area under the curve (AUC) using the receiver operating characteristic (ROC) curve to assess the flare classification performance.

The paper is structured as follows: Section 2 describes the solar flare data set. Section 3 outlines the proposed methods, including regularization of solar flare time series, sliding window, resampling of windows, LSTM and DLSTM algorithms, bagging-based ensemble algorithm, and performance metrics. Sections 4 and 5 present results and discussions, respectively. Section 6 summarizes the concluding remarks.

**Table 1**
X-Ray Flux for A-, B-, C-, M-, and X-class Flares

| Flare Class | X-Ray Flux Range (W m$^{-2}$) |
| --- | --- |
| A | $[10^{-8}, 10^{-7})$ |
| B | $[10^{-7}, 10^{-6})$ |
| C | $[10^{-6}, 10^{-5})$ |
| M | $[10^{-5}, 10^{-4})$ |
| X | $[10^{-4}, \infty)$ |

## 2. Data Set

The GOES, operated by NOAA, have monitored the Sun's X-ray flux and particle emissions since 1975. These satellites provide data on X-ray flux for various solar flaring events over different time intervals. In this study, we utilized a catalog of solar flares based on soft X-ray observations from GOES, covering the years 2003 November 1–2023 April 4 (N. Plutino et al. 2023). The catalog includes a total of 2800, 85,724, 59,995, 2382, and 170 flares of A, B, C, M, and X class, respectively. Table 1 presents the classification of A-, B-, C-, M-, and X-class solar flares based on their X-ray flux levels, as measured by GOES in the 1–8 Å wavelength band.

Each entry in the catalog contains details such as event ID, start time, end time, peak time, flare class, background flux, multiple ID, and total flux. Identifying more minor flares, such as A-class flares, is particularly challenging due to contamination from background solar X-ray radiation, leading to their lower reported numbers compared to B-class flares. On the other hand, M- and X-class flares, which are highly energetic, occur much less frequently than the medium-sized B- and C-class flares. This distribution aligns with the self-organized criticality behavior of the Sun's atmosphere (A. Gheibi et al. 2017).

## 3. Method

### 3.1. Regularization of Time Series for Solar Flare Prediction

Due to the self-organized criticality behavior, solar flare time series often exhibit high variability, noise, and non-stationarity, which poses significant challenges for reliable predictive modeling (M. Aschwanden 2011). Regularization techniques aim to enhance data consistency, reduce complexity, and improve signal quality, making models more reliable. In solar flare prediction, such preprocessing is crucial for addressing irregular sampling and filtering out irrelevant fluctuations, enabling machine learning algorithms to capture significant patterns of solar activity better (C. Bishop 2006).

A key regularization step in this study is resampling. We adopt a fixed 3 hr interval, converting irregular flare records into a uniform time series, thereby mitigating issues from missing data and variable cadences that could bias the model. We also apply filtering to eliminate long inactive periods, which enhances the signal-to-noise ratio by focusing on flare-active intervals.

To build the regularized time series, we initially exclude days with no recorded flaring activity (i.e., no flares of any class listed in the catalog), thereby focusing the data set on relevant active periods. Short gaps (less than 21 hr) are filled with a minimum flux value of approximately $2.071 \times 10^{-8}$ W m$^{-2}$, which approximates the GOES background level and avoids artificial discontinuities. For gaps shorter than 3 hr, if a flare occurred at either end, the missing data is filled with the corresponding start or end flux, preserving temporal variability without introducing artifacts.

Within each 3 hr interval, we extract the maximum flare peak flux, providing a compact and informative input for training deep learning models such as LSTM and DLSTM. The regularization process, combined with sliding window techniques, improves model performance, particularly for rare but important large flares. Among the tested intervals, the 3 hr setting struck a balance by maintaining flare structure while avoiding oversmoothing or high-frequency noise. Longer intervals risk losing critical flare information, while shorter ones show performance similar to the raw irregular data.

### 3.2. Flare Sliding Window

Solar flare data is characterized by sequences of flares with varying flux levels and waiting times, forming an irregular time series (N. Farhang et al. 2019, 2022; X. Liu et al. 2023). To examine these sequences, we create sliding windows that (1) capture both peak flux values and waiting times in the irregular time series and (2) reveal quasi-patterns of peak fluxes within the regularized time series. In the study of solar flares, waiting time represents the duration between successive flaring events. It plays a crucial role in analyzing the statistical characteristics of solar activity and the underlying energy release processes. Each sliding window consists of two subwindows of equal length: one for peak fluxes (peak flux subsliding window) and the other for waiting times (waiting time subsliding window). To improve the performance of learning algorithms, the flare peak fluxes and waiting times are scaled to a range between 0 and 1 by using their respective minimum and maximum values. The peak flux subsliding window contains 24 consecutive peak flux values, which may represent a quasi-pattern for flare activity. Similarly, the waiting time subsliding window captures 24 consecutive waiting times, reflecting a quasi-pattern of temporal intervals in the flare sequence. Using these sliding windows, we aim to identify diverse patterns in peak fluxes and waiting times, which are critical for forecasting flare activity based on the underlying time series.

Figure 1 illustrates an example of a sliding window with a length of $w = 24$, along with its corresponding label. In this setup, the first sliding window $w_1$ (green rectangle) includes the first 24 peak flux values in the peak flux subwindow (red rectangle) and the corresponding 24 waiting times in the waiting times subwindow (blue rectangle). The label for $w_1$ is assigned based on the class of the 25th flare. Flare classes are numerically encoded as 1, 2, 3, 4, and 5 for A-, B-, C-, M-, and X-class flares, respectively. This labeling approach ensures that each sliding window is linked to the subsequent flare's classification, facilitating the development of predictive models. This window illustrates an example of a quasi-pattern formed by a sequence of 24 peak fluxes and their corresponding waiting times in a solar flare time series.

Initially, we apply the sliding window algorithm to the solar flare time series, which contains a total of $N = 151,071$ flares. Using a window length of $w = 24$, the algorithm generates a total of $N - w = 151,047$ sliding windows, with each window capturing a sequence of $w$ consecutive flares. This process ensures that the entire time series is segmented into overlapping windows, enabling detailed analysis and pattern extraction.

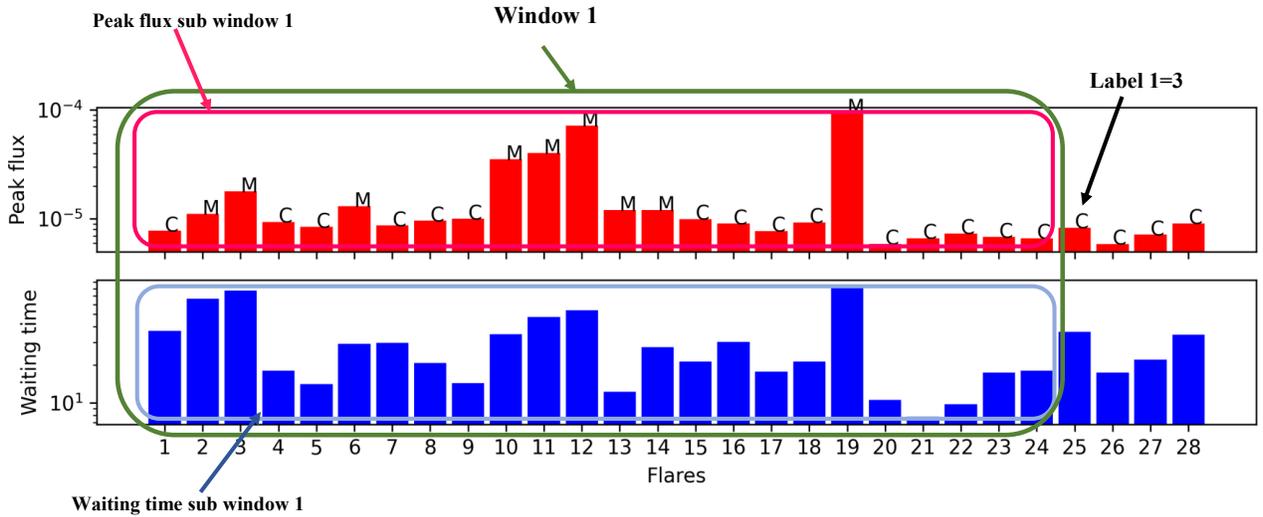

**Figure 1.** Example of sliding window $w_1$ (green rectangle) for 24 sequence flares, including the peak flux subwindow (red rectangle) and waiting time subwindow (blue rectangle). The label $1 = 3$ for $w_1$ indicates the 25th in the flare time series, which is C-class flares.

### 3.3. Resampling of Windows

The uneven distribution of flare classes poses a significant challenge in solar flare classification, framing it as an imbalanced classification problem (N. Alipour et al. 2019; X. Huang et al. 2024). To mitigate this issue, resampling methods are often employed to adjust the data set's class distribution, and various strategies have been proposed to enhance model performance under such conditions (N. Moniz et al. 2017; P. A. Vysakh & P. Mayank 2023).

This imbalance leads to differing performance of learning algorithms on the majority and minority classes, a phenomenon known as performance bias. Several solutions have been developed to tackle this issue effectively. W. Chen et al. (2024) categorized approaches into five groups: general methods, ensemble learning, imbalanced regression and clustering, long-tail learning, and imbalanced data streams. General methods encompass data-level techniques (such as oversampling and undersampling), algorithm-level strategies, and hybrid models. In our work, we apply oversampling to increase the representation of minority-class windows, under-sample the majority class for regularization, and use ensemble learning to enhance solar flare prediction.

Using a sliding window technique, we generated 151,047 labeled data segments. Our analysis revealed a significant imbalance among the five flare classes. To address this, we oversampled the data set. Flares were categorized into small (A-, B-, C-class) and large (M- and X-class) groups, yielding 148,519 small flare windows and just 2552 large flare windows- a ratio of approximately 58:1. To compensate, we duplicated the M- and X-class samples $R$ times, where $R$ was optimized based on classifier performance metrics such as the TSS.

### 3.4. LSTM Neural Networks

LSTM is a variant of recurrent neural network (RNN) created to resolve the vanishing gradient issue that affects conventional RNNs (S. Hochreiter & J. Schmidhuber 1997). RNNs are designed to process sequential data and retain information from previous iterations, which is useful for learning long-term dependencies. LSTM improves on this by selectively retaining only the useful information in time series forecasting, using a memory cell. An LSTM memory cell comprises three gates: the input, output, and forget gates. The core structure of an LSTM model consists of three interconnected layers: the input, hidden, and output layers (C. Wang et al. 2023; G. Jerse & A. Marcucci 2024; D. Salman et al. 2024).

The input gate regulates the input and previous hidden state, updating only the necessary information for accurate predictions. The forget gate uses a sigmoid activation function to discard irrelevant information from the memory cell. The output gate determines which data from the memory cell is transferred to the next hidden unit. These processes enable LSTM to learn and retain sequential dependencies effectively. In this study, we employ deep LSTM networks to identify patterns in multivariate time series.

### 3.5. Decomposition

Time series decomposition is a statistical approach in time series analysis that separates data into fundamental components, each representing different quasi-patterns (Q. Chen et al. 2019; G. Zhang et al. 2024). This method helps uncover historical variations within a time series and serves as a preprocessing step for forecasting models like Autoformer (H. Wu et al. 2021), Prophet (S. J. Taylor & B. Letham 2018), and DeepGLO (R. Sen et al. 2019). In deep learning models, decomposition functions as an internal mechanism, isolating hidden time series components from the main sequence. It divides the series into trend-cyclical and seasonal elements, capturing long-term developments and recurring patterns, respectively. To reduce periodic fluctuations and highlight long-term trends, the moving-average technique is utilized within the decomposition process.

$$X_t = \mathrm{AvgPool}(\mathrm{Padding}(X)), \quad (1)$$

$$X_s = X - X_t. \quad (2)$$

Here, $X_t$ and $X_s$ represent the trend-cyclical (Equation (1)) and seasonal (Equation (2)) components, respectively. The moving average is implemented using the AvgPool(.) operation with padding to preserve the original series length (H. Wu et al. 2021).

### 3.6. Bagging-based Ensemble Algorithm

Predicting solar flares is challenging due to the complex and dynamic nature of solar activity. In this study, we apply the bagging (bootstrap aggregation) method to enhance the performance of LSTM and DLSTM models for solar flare forecasting. Bagging is an ensemble learning technique that improves predictive performance by training multiple models on different subsets of the data set and then averaging their predictions. This approach helps reduce overfitting, increases stability, and enhances the generalization capability of machine learning models (R. E. Schapire 1990; L. Breiman 1996; T. G. Dietterich 2000; M. Mngomezulu et al. 2023; A. Mohammed & R. Kora 2023).

In our implementation, we apply bagging to both LSTM and DLSTM models using a regularized solar flare time series, where the data is segmented into fixed 3 hr intervals. Each LSTM and DLSTM model is trained on a bootstrapped subset of the data, and their outputs are averaged to produce a more robust prediction.

### 3.7. Performance Metrics

The confusion matrix is a fundamental tool for evaluating classification models, providing insight into the number of correct and incorrect predictions. It is composed of four elements: true positives (TP), the number of positive instances correctly predicted as positive; false positives (FP), the number of negative instances incorrectly predicted as positive; true negatives (TN), the number of negative instances correctly predicted as negative; and false negatives (FN), the number of positive instances incorrectly predicted as negative. These elements are used to calculate several performance metrics that evaluate the model's effectiveness. Please note that the sizes of positive and negative classes are P = TP + FN and N = FP + TN, respectively.

Accuracy (predictive value) measures the fraction of correctly identified positive cases among all predicted positives, while recall (sensitivity, true positive rate (TPR)) reflects the proportion of actual positive cases that are correctly predicted. However, in highly imbalanced data sets such as solar flare forecasting, where small flares vastly outnumber large ones—these metrics can be misleading. A model may achieve high accuracy, precision, and ApSS by mainly predicting the dominant class (e.g., small flares), yet fail to detect rare, impactful large flares.

In the context of highly imbalanced classification problems like solar flare prediction—where small flares vastly outnumber large, more impactful ones—the choice of performance metric is critical (C. A. Doswell et al. 1990; D. S. Bloomfield et al. 2012). Although the Heidke skill score (HSS) has been widely used, it suffers from a significant limitation: it is sensitive to class imbalance and often inflates the model's performance when biased toward the majority class. This is particularly problematic in solar flare forecasting, where the true value lies in correctly identifying rare but important large flares (L. D. Krista 2025). HSS evaluates prediction skill relative to random chance, but does not sufficiently penalize models that fail to detect minority-class

**Table 2**
Performance Metrics (Accuracy, Precision, Recall, ApSS, HSS, and TSS) for a Binary Classifier Related to the Elements of the Confusion Matrix

| Metric | Formula |
| --- | --- |
| Accuracy | $\text{Accuracy} = \frac{\text{TP} + \text{TN}}{\text{TP} + \text{FP} + \text{TN} + \text{FN}}$ |
| Precision (positive predictive value) | $\text{Precision} = \frac{\text{TP}}{\text{TP} + \text{FP}}$ |
| Recall (sensitivity or TPR) | $\text{Recall} = \frac{\text{TP}}{\text{TP} + \text{FN}}$ |
| Specificity (TN rate) | $\text{Specificity} = \frac{\text{TN}}{\text{TN} + \text{FP}}$ |
| ApSS | $\text{ApSS} = \begin{cases} \frac{\text{TP} - \text{FP}}{\text{TP} + \text{FN}} & \text{if TP} + \text{FN} < \text{TN} + \text{FP} \\ \frac{\text{TN} - \text{FN}}{\text{TN} + \text{FP}} & \text{if TP} + \text{FN} \geqslant \text{TN} + \text{FP} \end{cases}$ |
| HSS | $\text{HSS} = \frac{2 \times [\text{TP} \times \text{TN} - \text{FN} \times \text{FP}]}{P \times (\text{TN} + \text{FN}) + (\text{TP} + \text{FP}) \times N}$ |
| TSS | $\text{TSS} = \text{sensitivity} + \text{specificity} - 1$ |

events, leading to potentially misleading assessments of predictive capability.

In contrast, the TSS, also known as the Hanssen and Kuipers discriminant, offers a more robust and unbiased evaluation. Defined as TSS = sensitivity + specificity − 1, it treats both classes symmetrically and remains unaffected by class proportions. TSS provides a clear picture of a model's ability to distinguish between flare and no-flare (or small and large flare) events across all thresholds, making it especially suitable for imbalanced data sets. Leading researchers, including D. S. Bloomfield et al. (2012), have recommended TSS as the standard metric when comparing forecasting models with varying flare/no-flare ratios. Therefore, in studies aiming to accurately predict large solar flares—a minority class of critical importance—TSS should be prioritized over HSS or accuracy-based metrics, as they fail to reflect true forecasting skill under severe class imbalance (D. S. Bloomfield et al. 2012; M. G. Bobra & S. Couvidat 2015; N. Alipour et al. 2019).

To examine how HSS and TSS behave when the data is very imbalanced, let us consider a simple binary classification case where the model correctly predicts 75% of the positive class (TP = 0.75P) and misses 25% (FN = 0.25P), while also correctly identifying 75% of the negative class (TN = 0.75N) and incorrectly classifying 25% as positive (FP = 0.25N). Using these values and the formula for HSS from Table 2, we get

$$\text{HSS} = \frac{4\frac{P}{N}}{\left(\frac{P}{N} + 3\right)^2 - 8}. \quad (3)$$

As shown in Equation (3), when the data set is heavily imbalanced ($N \ll P$ or $P \ll N$), HSS becomes very small. In contrast, TSS for this confusion matrix is calculated as 0.50, which gives a more reasonable sense of the model's ability to separate the two classes. This shows that HSS is very sensitive to class imbalance, while TSS is not.

Table 2 represents the key metrics, their formulas, and their relationships to the confusion matrix components. This table provides a clear overview of how each metric is calculated and its relation to the elements of the confusion matrix. These metrics, particularly TSS, are crucial for assessing the performance of models in binary classification tasks, especially when dealing with imbalanced data.

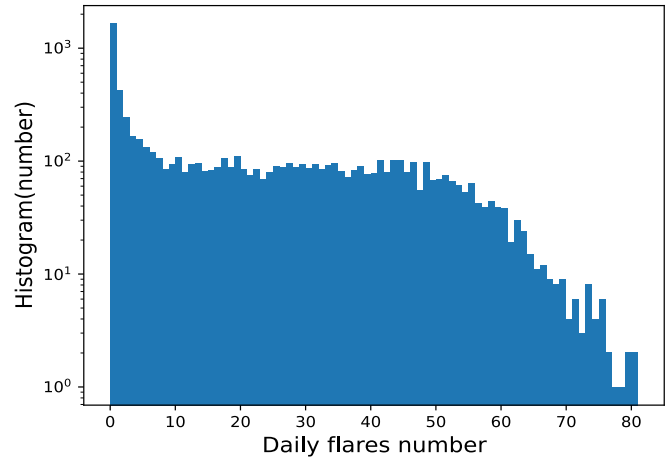

**Figure 2.** The frequency size distribution (histogram) of daily flare number for 2003 January 11 to 2023 April 30.

## 4. Results

We apply LSTM and DLSTM networks for the sliding window pattern to forecast solar flares in the irregular (original data) and regular flare time series. First, we use the algorithm to smooth the peak values of flares in the catalog containing 151,071 flares. Our data catalog includes X-ray GOES flux for flares in the $2.071 \times 10^{-08}$ W m$^{-2}$ and $2.628 \times 10^{-3}$ W m$^{-2}$. We replace the flux value with the corresponding bin's mean value in the smoothed time series. We use the 500 equal spacing binning for the flux to do this. Smoothing flare time series can significantly impact the application of machine learning models for forecasting flares due to noise reduction (K. Ozawa et al. 2024; F. Shan et al. 2024), trend identification (A. P. Wibawa et al. 2022), and overfitting reduction (L. Li & M. Spratling 2023) in the time series. The percentage of flares A-, B-, C-, M-, and X-class flares in our data set is 1.85, 56.75, 39.71, 1.58, and 0.11, respectively, which implies that the flares are imbalanced class problems. We launch a GitHub[4] project that provides the solar flare data set, along with Python notebooks to implement LSTM and DLSTM models for flare forecasting.

### 4.1. Statistics Quasi-patterns in Solar Flares

We apply the sliding window approach to recognize the quasi-patterns in the irregular flares time series. Since the flares result in stochastic changes in the structures and topology of the magnetic fields of the complex Sun, we expect the complex pattern in the flares time series within the solar cycles (S. Taran et al. 2022; Z. Tajik et al. 2024). However, similar to sunspots, the system of flares shows the butterfly diagram representing the flares' distribution within the solar surface and the 11 yr cycle (A. Gheibi et al. 2017).

To determine the optimal window size, we examined the frequency distribution (histogram) of the daily flare counts. Figure 2 presents this distribution for data spanning from 2003 November 1 to 2023 April 4. Over these 20 yr of X-ray flare observations, we identified 1646 days without recorded flares, with an average of 20.4 flares per day. The maximum daily flare count was 81, recorded on 2014 May 2. The flaring activity follows a Poisson process (M. S. Wheatland & Y. E. Litvinenko 2002; N. Farhang et al. 2019), and the average daily flare count (20 flares day$^{-1}$) is a reasonable indicator for selecting quasi-pattern windows.

Based on this, we chose a window size of $w = 24$, representing 24 consecutive flaring events, for the sliding window algorithm. This size slightly exceeds the average daily flare count, ensuring sufficient quasi-pattern representation within each window. Each window comprises sequences of 24 peak flux values and waiting times, effectively capturing flare time series patterns. Although longer windows could improve performance in machine learning algorithms for flare prediction, they significantly increase computational costs with only marginal gains. On the other hand, shorter windows fail to capture adequate patterns and trends in flux and waiting times, leading to suboptimal results. Therefore, a window length of 24 provides an efficient balance, capturing necessary patterns effectively. R. Tang et al. (2021) previously applied sliding window techniques to active region magnetic parameters for solar flare prediction. L. F. L. Grim & A. L. S. Gradvohl (2024) used a similar approach with SHARP magnetogram sequences to forecast significant M- and X-class flares.

The sliding window algorithm generates $N − w$ pattern windows of length $w$ from a time series of length $n$. For $N = 151{,}071$ and $w = 24$, this results in 151,047 pattern windows, averaging 7552 windows per year over the data set's 20 yr span. Considering four flare classes of A, B, C, and large (M and X) classes, the potential pattern combinations for $w = 20$ are approximately $4^{20} \approx 10^{12}$, vastly outnumbering the observed 151,047 windows. This discrepancy underscores the vast complexity of the solar flare system and its limited predictability over extended periods. These characteristics, along with the system's nonlinearity and self-organized criticality, align with the features of complex systems (Y. Bar-Yam 1997; J. Crutchfield & K. Young 1988; R. Foote 2007; R. S. MacKay 2008; A. Gheibi et al. 2017).

Figure 3 illustrates the probability density function (PDF) for the minimum (left panel) and maximum (right panel) peak fluxes within the analyzed windows. The red dashed lines represent the power-law function fitted to the data, characterized by $\alpha = 3.27$ and $f_{\rm th} = 7.76 \times 10^{-7}$ W m$^{-2}$ for the minimum fluxes, and $\alpha = 1.89$ and $f_{\rm th} = 5.63 \times 10^{-6}$ W m$^{-2}$ for the maximum fluxes. The power-law model is mathematically

---
[4] https://github.com/ZeinabHassani/SolarFlarePredition

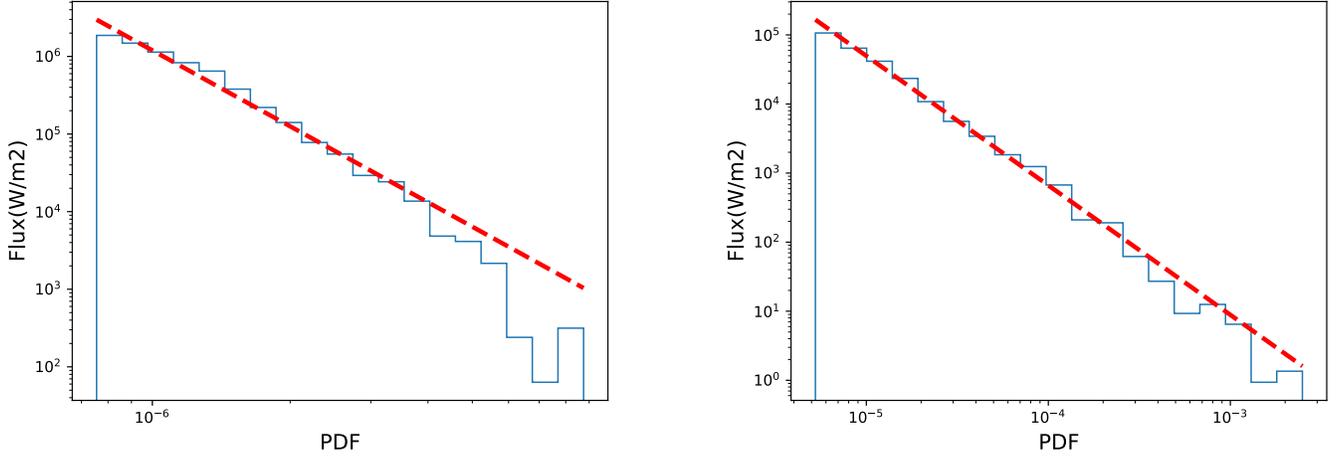

**Figure 3.** PDF for the minimum (left panel) and maximum (right panel) for windows of peak fluxes. The red dashed line represents the power-law function, where for the minimum flux of the windows, $\alpha = 3.27$ and $f_{\text{th}} = 7.76 \times 10^{-7}$ W m$^{-2}$; for the maximum flux of the windows, $\alpha = 1.89$ and $f_{\text{th}} = 5.63 \times 10^{-6}$ W m$^{-2}$.

expressed as

$$p(f, f_{\text{th}}, \alpha) = \frac{\alpha - 1}{f_{\text{th}}} \left( \frac{f}{f_{\text{th}}} \right)^{-\alpha}, \quad (4)$$

where $f$ denotes the flux value, $f_{\text{th}}$ is the flux threshold, and $\alpha$ is the power-law index. The parameter $\alpha$ in Equation (4) is estimated using the maximum likelihood estimation method (N. Farhang et al. 2018, 2019), while $f_{\text{th}}$ is determined via the Kolmogorov–Smirnov test. As depicted in Figure 3, both the minimum and maximum fluxes within the peak flux windows exhibit a power-law distribution, implying that the flare fluxes display characteristics consistent with a self-organized criticality system. This power-law behavior underscores the principle that self-organized criticality is an intrinsic property of the flare system (M. K. Georgoulis et al. 2001; M. Aschwanden 2011; A. Strugarek & P. Charbonneau 2014; N. Farhang et al. 2022).

To achieve high-performance flare prediction, it is essential to implement techniques that effectively address class imbalance in flare time series data. One effective approach to mitigating this issue is regularizing the time series, ensuring a more uniform distribution of data points.

Since solar flare time series data exhibit high irregularity, we first remove all days without flares from the data set. This step eliminates unnecessary gaps, ensuring that the model focuses on meaningful flare activity rather than extended periods of inactivity. Next, we divide the remaining time series into 3 hr intervals, selecting the largest peak within each interval to create a more structured data set. The choice of a 3 hr interval for regularization is an ad hoc decision, balancing the need to preserve structural information while avoiding excessive fragmentation. Longer intervals (e.g., greater than 4 hr) may smooth out essential flare patterns, leading to a loss of critical structural details.

Conversely, shorter intervals yield results that slight resemble the irregular time series, offering little improvement. After testing different time intervals, we found that a 3 hr window provides the optimal balance, achieving the highest performance for our data set.

By applying this regularization process, we transform the data set into a structured time series containing 47,152 data points, where each flare-active day is represented by eight flux values recorded at 3 hr intervals. Although the data set remains imbalanced, this approach reduces the imbalance by a factor of approximately 2.5, which enhances the flare-forecasting performance, particularly for larger flares. We construct sliding windows for the regular flare flux time series while excluding waiting times from the window structure. Instead, we use a flux-only subwindow, which helps the model capture flare dynamics more effectively.

### 4.2. LSTM and DLSTM Forecasting Models

We apply LSTM and DLSTM networks to predict two classes of flares: small flares (A, B, and C classes) and large flares (M and X classes). The LSTM model architecture includes four LSTM layers for time series processing. The activation functions used are the rectified linear unit (ReLU) and tanh, followed by a fully connected layer with a sigmoid activation function. The ReLU, sigmoid, and hyperbolic tangent activation functions are widely used nonlinear functions in machine learning and neural networks. The ReLU function, defined as it has become the state-of-the-art activation function due to its simplicity and superior performance. The ReLU function is given by

$$\text{ReLU}(x) = x^+ = \max(0, x) = \frac{x + |x|}{2} = \begin{cases} x & \text{if } x > 0, \\ 0 & \text{if } x \leqslant 0. \end{cases} \quad (5)$$

In the neural network, the use of the hyperbolic tangent and sigmoid functions was inspired by the firing behavior of biological neurons (X. Chen et al. 2006; Y. LeCun et al. 2015). The hyperbolic tangent and sigmoid functions are defined by

$$\tan h(x) = \frac{e^{2x} - 1}{e^{2x} + 1}, \quad (6)$$

$$s(x) = (1 + e^{-x})^{-1}. \quad (7)$$

The decomposition model (DLSTM) incorporates a decomposition block and a fully connected layer for the time series, one LSTM layer, and two fully connected layers. To classify the sliding windows, including the peak flux and waiting time subwindows, the model is trained with *epochs* = 30 and *batchsize* = 128. The training data set comprises 80% of the sample windows, selected from the initial portion of the time series, while the test set consists of the remaining 20%, chosen from the final portion of the time series. During training, we employ fourfold cross-validation ($k = 4$) to validate the model. Due to the temporal dependencies inherent in time series data, cross validation for machine learning analyses must ensure that the training set consists of earlier data and the test set is drawn from later data. This approach helps to avoid information leakage and ensures meaningful temporal validation (N. Nishizuka et al. 2021; J. Eiglsperger et al. 2023).

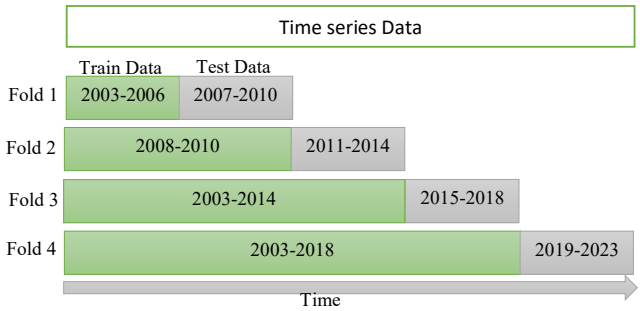

**Figure 4.** The solar flare splitting scheme for regularized time series cross-validation of Folds 1–4. For each fold, the green box represents the time span used for the training data set, while the gray box corresponds to the testing data set duration.

Figure 4 illustrates the cross-validation scheme applied to the regularized solar flare data set. In this setup, each fold uses 20% of the data for testing. Specifically, Fold 1 uses data from 2003 to 2006 for training and from 2006 to 2010 for testing. Fold 2 expands the training period to 2003–2010 and tests on 2011–2014. Fold 3 trains from 2003–2014 and tests from 2015 to 2018. Finally, Fold 4 uses 2003–2018 as the training interval and tests on the 2019–2023 data.

Figure 5 illustrates the process by which DLSTM effectively extracts trends (red dashed line) and seasonal (blue line) components from the irregular flare time series (black line) over 200 time steps (top panel). The bottom panel displays the random noise, derived by subtracting the sum of the trends and seasonal components from the original flare time series. The DLSTM achieves this by leveraging a decomposition-based approach that separates the meaningful patterns in the data (trends and seasonal variations) from random, irregular fluctuations. Using advanced smoothing algorithms, the model carefully isolates the underlying trends representing long-term behavior and the periodic seasonal patterns inherent in the time series.

The smoothing function is based on the moving-average method, which helps reduce short-term fluctuations by averaging values over a defined window. The moving average is a widely used smoothing technique in time series analysis, defined as the arithmetic mean of a set of sequential data points within a fixed-size window, effectively reducing noise while preserving significant patterns. These components are then utilized as inputs to the model for accurate prediction, while the random noise is discarded.

As illustrated in Figure 5, the DLSTM effectively captures the trend and seasonal components, with their combined representation closely matching the original flare time series, leaving only minimal random noise. The minimal random noise reflects the robustness and validity of the smoothing algorithm employed in this work, demonstrating its effectiveness in isolating significant patterns while removing unnecessary variability. This decomposition approach not only ensures better model performance but also highlights the importance of preprocessing steps in time series forecasting tasks (H. Wu et al. 2021; B. Wu et al. 2023; E. Estrada-Patiño et al. 2024; K. Zhang et al. 2024).

Ensemble models are machine learning techniques that combine the predictions of multiple models, enhancing overall performance, accuracy, and robustness compared to using a single model. This approach leverages the idea that aggregating diverse models helps compensate for individual model weaknesses, ultimately reducing prediction errors and improving reliability. Hereafter, we focus on analyzing the outputs of six machine learning algorithms, including (1) LSTM applied to irregular time series, (2) DLSTM on irregular time series, (3) LSTM on regularized time series, (4) DLSTM on regularized time series, (5) LSTM on regularized time series incorporating ensemble learning, and (6) DLSTM on regularized time series with ensemble learning.

Since solar flares originate from the Sun's cyclic magnetic activity, their intensity is expected to be influenced by this periodic behavior. To emphasize the impact of solar activity on flare-forecasting methods, we analyze the performance of LSTM and DLSTM algorithms on irregular time series.

The TPR, or sensitivity, measures the proportion of large solar flares (M, X) accurately identified by the model, while the false positive rate (FPR) quantifies the proportion of small flares (A, B, C) incorrectly classified as large. These metrics are essential for evaluating model performance in binary classification tasks, balancing effective detection of significant flares with minimizing false alarms. TPR and FPR are computed for classifying small (A, B, C) and large (M, X) solar flares.

Figure 6 illustrates the average TSS, along with standard deviation error bars, for LSTM (blue circles) and DLSTM (red triangles) across different K-folds of the binary forecasting model for small and large solar flares, with resampling set to $R = 12$ for the minority class. The TSS measures the balance between the TPR (sensitivity) and the FPR (specificity) in handling highly imbalanced solar flare data.

For K-fold = 1, both LSTM and DLSTM used the first quarter of the flare time series (2003–2012), which includes windows of peak fluxes and waiting times. The training set comprised the first 80 of the time series, while the test set used the last 20. The resulting TSS values were $0.42 \pm 0.15$ for LSTM and $0.51 \pm 0.04$ for DLSTM. For K-fold = 2, the time series spanned 50 of the data (2003–2014), covering the declining phase of solar cycle 23 (2003–2009) and the rising phase of cycle 24 (2009–2012). Despite having twice the number of pattern windows as K-fold = 1, the TSS performance slightly decreased.

This discrepancy may be due to the fact that K-fold = 1 focused solely on patterns from the declining phase of cycle 23 and the rising phase of cycle 24, whereas K-fold = 2 encompassed patterns from the declining, rising, and peak phases. The training data for K-fold = 2 (2003–2012) included only limited patterns from the declining phase of cycle 23 and the rising phase of cycle 24, making it challenging to accurately recognize patterns during the test period (2012–2014), which corresponds to the peak phase (maximum activity) of cycle 24. This highlights the complexities of solar cycles and the associated variations in sunspot and flare activities. For K-fold = 3 and K-fold = 4, which include more comprehensive flare pattern windows from cycles 24 and 25, TSS performance improved slightly for both LSTM and DLSTM. Notably, K-fold = 4 spans 20 yr of solar flare time series data, providing a balanced representation of increasing and decreasing phases, further enhancing model performance. These findings highlight the importance of incorporating diverse solar activity patterns for robust flare forecasting for both models. Also, we obtain a similar result for LSTM and DLSTM for regular flare time series.

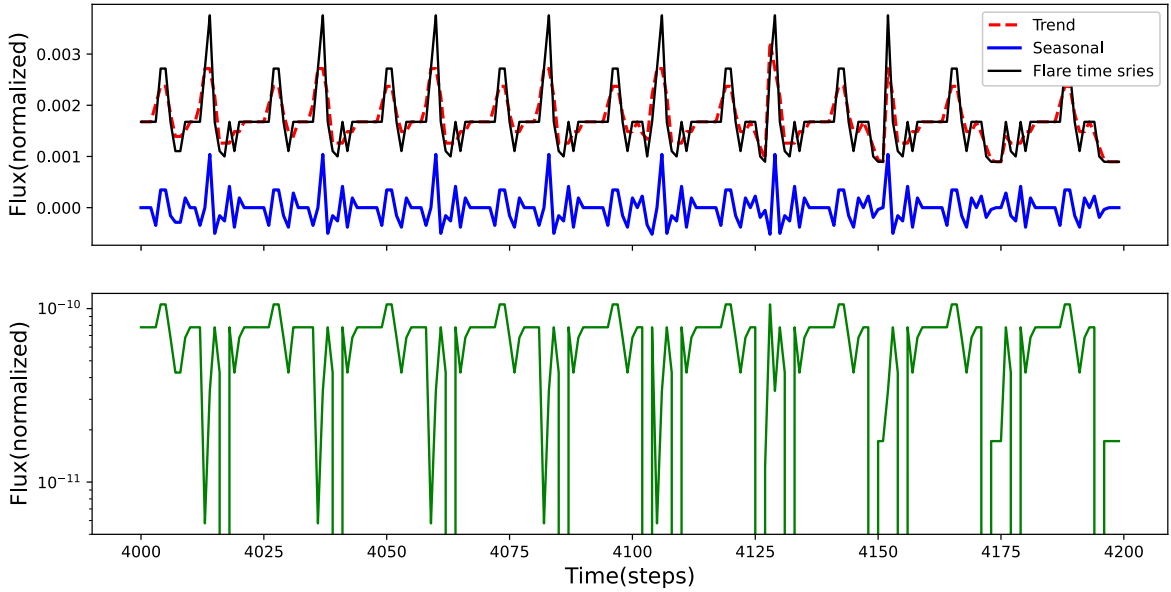

**Figure 5.** (Top panel) The DLSTM extracts trends (red dashed line) and seasonal components (blue line) from the flare time series (black line) over 200 time steps. (Bottom panel) The random noise is derived by subtracting the sum of the trends and seasonal components from the flare time series.

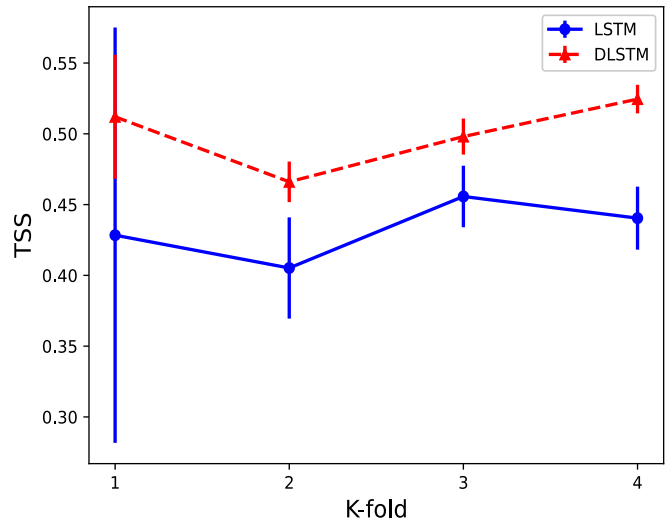

**Figure 6.** The average to gather with error bar (standard deviation) TSS for LSTM (blue circle) and DLSTM (red triangle) for different K-folds of the binary forecasting model for small and large flares with resampling $R = 12$ of the minority class.

Figure 7 presents the ROC curves for six models: LSTM on irregular time series (Model 1, blue line), DLSTM on irregular time series (Model 2, orange line), LSTM on regular time series (Model 3, green line), DLSTM on regular time series (Model 4, red line), LSTM on regular time series with ensemble learning (Model 5, purple line), and DLSTM on regular time series with ensemble learning (Model 6, brown line). The left panel shows results without resampling ($R = 0$), while the right panel illustrates results with resampling ($R = 12$). A green dashed line represents a random classifier. The ROC curve plots the TPR against the FPR across different thresholds. The AUC values obtained for the models are 0.50 (0.68), 0.59 (0.77), 0.55 (0.84), 0.53 (0.84), 0.67 (0.75), and 0.65 (0.87) for $R = 0$ ($R = 12$), respectively. These correspond to LSTM on irregular time series, DLSTM on irregular time series, LSTM on regular time series, DLSTM on regular time series, LSTM on regular time series with ensemble learning, and DLSTM on regular time series with ensemble learning.

All six models perform slightly better than a random classifier (AUC = 0.5) when no resampling is applied ($R = 0$). However, with resampling ($R = 12$), their ability to forecast both large and small flares improves significantly. Notably, DLSTM on regular time series with the ensemble approach achieves the highest performance. Interestingly, LSTM and DLSTM applied to regular time series outperform their counterparts on irregular time series. This suggests that using a regular time interval reduces complexity and noise compared to irregular time series.

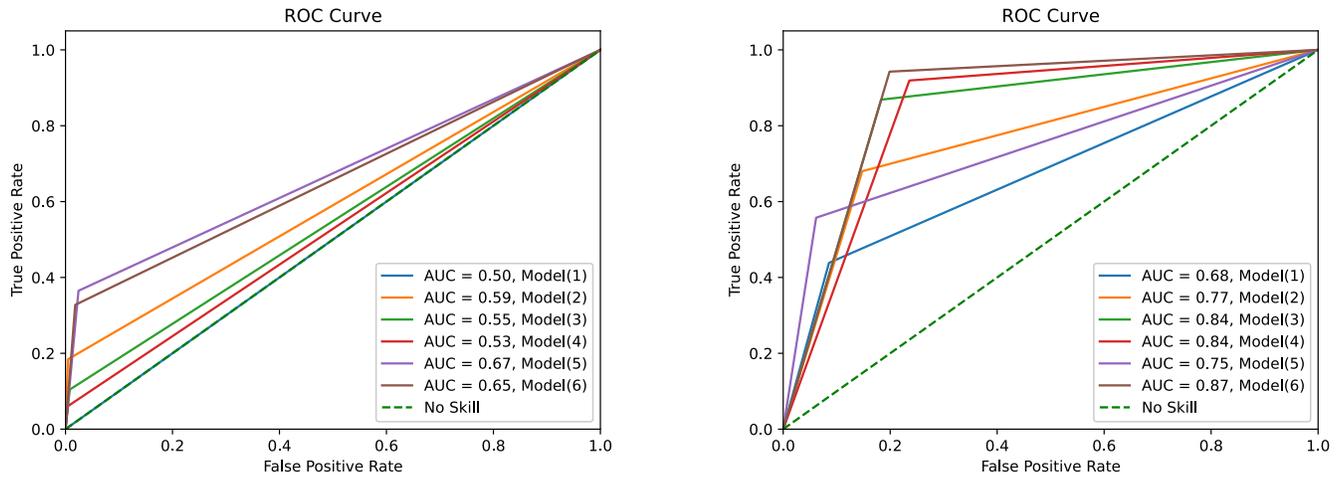

**Figure 7.** A sample of the ROC curves for six models: LSTM on irregular time series (Model 1), DLSTM on irregular time series (Model 2), LSTM on regular time series (Model 3), DLSTM on regular time series (Model 4), LSTM on regular time series with ensemble learning (Model 5), and DLSTM on regular time series with ensemble learning (Model 6). The left panel shows results without resampling ($R = 0$), while the right panel includes resampling ($R = 12$). The green dashed line represents a random classifier with no predictive skill. The AUC is calculated for each model, where an AUC of 0.5 indicates performance of random classification.

The superior performance of DLSTM is attributed to its architecture, which decomposes the flare time series into trends, seasonal, and random noise components. Consequently, the decomposition-based approach of DLSTM offers a significant advantage in solar flare forecasting. By using trends and seasonal components in its fully connected layers, DLSTM effectively excludes random noise, enabling better predictive accuracy than LSTM (H. Wu et al. 2021; E. Estrada-Patiño et al. 2024).

It is important to note that AUC values may fluctuate slightly across iterations due to the random sampling of training and testing sets. To ensure a more reliable comparison of model performance across different sampling rates, computing the average and standard deviation over multiple iterations is essential. Additionally, analyzing various performance metrics for both positive and negative classes, along with global metrics, provides a more comprehensive evaluation of the forecasting classifier, particularly in cases of class imbalance. To address this, we obtain the actual values and standard deviations for performance metrics and AUC for each model by applying numerous random sampling iterations.

Table 3 summarizes the results of six models trained for solar flare prediction, each differing in time series format (irregular versus regular), resampling techniques, and LSTM architecture (LSTM versus DLSTM). The models utilize a sliding window approach on flare time series and include (1) LSTM applied to irregular time series, (2) DLSTM on irregular time series, (3) LSTM on regularized time series, (4) DLSTM on regularized time series, (5) LSTM on regularized time series incorporating ensemble learning, and (6) DLSTM on regularized time series with ensemble learning. Given the stochastic nature of LSTM and DLSTM neural networks, each model is trained over multiple iterations, and the average and standard deviation of the evaluation metrics are calculated to ensure robustness.

The models are evaluated using several key metrics: accuracy, recall, precision, TSS, ApSS, HSS, and AUC. LSTM applied to irregular time series demonstrates strong accuracy (0.98), but the recall for large flares is low (0.12) due to class imbalance. Resampling improves the recall for large flares (0.12–0.51), helping balance the detection of both small and large flares. AUC stays high (0.71), but precision for large flares remains relatively low. DLSTM on irregular time series shows slightly better recall for large flares (0.80) than LSTM, with accuracy (0.74). However, large flare precision still struggles. AUC improves to 0.77, making it the best-performing irregular time series model. For all six models, the HSS values remain relatively low. Although we applied slight oversampling to the minority class, the HSS did not significantly improve, indicating that the score remains sensitive to class imbalance.

LSTM on regularized time series (3 hr interval) with different resampling achieves a large flare recall of 0.88 and slightly improved in accuracy (0.78). It balances small and large flare detection better than the irregular time series models. DLSTM on regular time series (3 hr interval) with different resampling further improves large flare recall (0.94) while maintaining accuracy (0.74). TSS and ApSS scores are higher, indicating better reliability. AUC (0.84) suggests this model generalizes well across different flare classes.

LSTM on regularized time series (3 hr interval) with different resampling and ensemble learning refines predictions using the data set's internal patterns. Large flare recall increases to 0.70, surpassing previous models, with improved precision and accuracy (0.9). AUC remains around 0.78, ensuring more stable performance. DLSTM on regularized time series (3 hr interval) with different resampling and ensemble learning is the top-performing model, with the highest large flare recall (0.97) and strong accuracy (0.77). It also achieves the highest TSS and ApSS scores, outperforming all other models. AUC reaches 0.87, confirming its strong predictive ability.

Increasing the resampling rate in all six models leads to slight decreases in accuracy and recall for small flares. However, it helps address the class imbalance by improving large flare recall. This pattern is consistent across regularization and ensemble approaches, which enhance the model's performance by better handling the imbalance of large flares. These improvements are reflected in the AUC in the ROC curve, with higher TPR compared to FPR. Additionally, TSS improves with resampling, regularization, and ensemble

**Table 3**
Performance Metrics

| Data | Model | R | Accuracy | TSS | Small Flares | | Large Flares | | ApSS | HSS | AUC |
|---|---|---|---|---|---|---|---|---|---|---|---|
| | | | | | Precision | Recall | Precision | Recall | | | |
| Irregular flare time series | LSTM on irregular | 0 | 0.98 ± 0.001 | 0.12 ± 0.08 | 0.987 ± 0.001 | 0.997 ± 0.003 | 0.25 + 0.2 | 0.12 ± 0.08 | 0.98 ± 0.004 | 0.16 ± 0.1 | 0.56 ± 0.06 |
| | | 12 | 0.90 ± 0.01 | 0.41 ± 0.03 | 0.992 ± 0.001 | 0.91 ± 0.01 | 0.07 ± 0.01 | 0.51 ± 0.04 | 0.90 ± 0.01 | 0.11 ± 0.02 | 0.71 ± 0.02 |
| | | 20 | 0.88 ± 0.02 | 0.33 ± 0.03 | 0.991 ± 0.001 | 0.88 ± 0.02 | 0.06 ± 0.01 | 0.45 ± 0.04 | 0.88 ± 0.02 | 0.07 ± 0.01 | 0.67 ± 0.01 |
| | DLSTM on irregular | 0 | 0.98 ± 0.001 | 0.17 ± 0.03 | 0.988 ± 0.001 | 0.995 ± 0.001 | 0.37 ± 0.03 | 0.18 ± 0.03 | 0.98 ± 0.01 | 0.23 ± 0.03 | 0.59 ± 0.02 |
| | | 12 | 0.84 ± 0.02 | 0.54 ± 0.01 | 0.994 ± 0.0003 | 0.84 ± 0.02 | 0.060 ± 0.004 | 0.70 ± 0.02 | 0.84 ± 0.02 | 0.09 ± 0.007 | 0.77 ± 0.003 |
| | | 20 | 0.74 ± 0.05 | 0.54 ± 0.01 | 0.996 ± 0.001 | 0.74 ± 0.05 | 0.04 ± 0.01 | 0.80 ± 0.04 | 0.73 ± 0.05 | 0.06 ± 0.01 | 0.77 ± 0.01 |
| Regular flare time series | LSTM on regular | 0 | 0.96 ± 0.003 | 0.14 ± 0.07 | 0.97 ± 0.003 | 0.99 ± 0.01 | 0.30 ± 0.05 | 0.15 ± 0.07 | 0.96 ± 0.003 | 0.18 ± 0.07 | 0.57 ± 0.03 |
| | | 12 | 0.79 ± 0.03 | 0.68 ± 0.02 | 0.996 ± 0.001 | 0.79 ± 0.03 | 0.12 ± 0.01 | 0.89 ± 0.03 | 0.79 ± 0.03 | 0.16 ± 0.02 | 0.84 ± 0.01 |
| | | 20 | 0.78 ± 0.03 | 0.66 ± 0.02 | 0.995 ± 0.001 | 0.78 ± 0.03 | 0.10 ± 0.01 | 0.88 ± 0.04 | 0.78 ± 0.03 | 0.14 ± 0.07 | 0.83 ± 0.01 |
| | DLSTM on regular | 0 | 0.96 ± 0.002 | 0.11 ± 0.04 | 0.97 ± 0.01 | 0.99 ± 0.003 | 0.29 ± 0.02 | 0.11 ± 0.05 | 0.96 ± 0.001 | 0.14 ± 0.05 | 0.55 ± 0.02 |
| | | 12 | 0.78 ± 0.03 | 0.68 ± 0.01 | 0.996 ± 0.001 | 0.78 ± 0.03 | 0.11 ± 0.01 | 0.90 ± 0.04 | 0.87 ± 0.02 | 0.15 ± 0.02 | 0.84 ± 0.01 |
| | | 20 | 0.74 ± 0.07 | 0.67 ± 0.04 | 0.998 ± 0.001 | 0.73 ± 0.06 | 0.09 ± 0.02 | 0.94 ± 0.03 | 0.73 ± 0.07 | 0.13 ± 0.03 | 0.84 ± 0.02 |
| | LSTM with ensemble on regular | 0 | 0.96 ± 0.002 | 0.32 ± 0.03 | 0.984 ± 0.001 | 0.98 ± 0.01 | 0.27 ± 0.03 | 0.35 ± 0.03 | 0.96 ± 0.001 | 0.28 ± 0.3 | 0.66 ± 0.02 |
| | | 12 | 0.91 ± 0.02 | 0.57 ± 0.1 | 0.991 ± 0.004 | 0.91 ± 0.03 | 0.15 ± 0.02 | 0.65 ± 0.1 | 0.91 ± 0.02 | 0.22 ± 0.01 | 0.78 ± 0.06 |
| | | 20 | 0.90 ± 0.04 | 0.61 ± 0.06 | 0.993 ± 0.002 | 0.91 ± 0.03 | 0.15 ± 0.02 | 0.70 ± 0.07 | 0.90 ± 0.03 | 0.22 ± 0.01 | 0.81 ± 0.04 |
| | DLSTM with ensemble on regular | 0 | 0.97 ± 0.01 | 0.21 ± 0.01 | 0.98 ± 0.003 | 0.99 ± 0.01 | 0.30 ± 0.1 | 0.23 ± 0.2 | 0.97 ± 0.01 | 0.22 ± 0.13 | 0.61 ± 0.07 |
| | | 12 | 0.80 ± 0.01 | 0.74 ± 0.003 | 0.998 ± 0.001 | 0.80 ± 0.01 | 0.097 ± 0.003 | 0.95 ± 0.01 | 0.80 ± 0.01 | 0.14 ± 0.005 | 0.87 ± 0.001 |
| | | 20 | 0.77 ± 0.01 | 0.73 ± 0.01 | 0.999 ± 0.0001 | 0.77 ± 0.01 | 0.084 ± 0.002 | 0.967 ± 0.004 | 0.75 ± 0.01 | 0.12 ± 0.005 | 0.87 ± 0.004 |

**Note.** The performance metrics, including accuracy, recall, precision, TSS, ApSS, HSS, and AUC, are calculated for (1) LSTM applied to irregular time series, (2) DLSTM on irregular time series, (3) LSTM on regularized time series, (4) DLSTM on regularized time series, (5) LSTM on Regularized time series incorporating ensemble learning, and (6) DLSTM on regularized time series and ensemble learning with resampling $R = 0$, $R = 12$, and $R = 20$. For each metric, we calculate the average (actual) value along with the standard deviation, which indicates the variability as ± values.

learning, helping identify the best model for forecasting large flares.

## 5. Comparison and Discussion

Solar flare prediction has seen a range of methodologies, each leveraging different types of data and machine learning techniques (N. Lotfi et al. 2020; K. D. Leka et al. 2023; X. Huang et al. 2024). Broadly, these methods fall into categories: those based on image-derived features, those using time series data sets, and those applying hybrid features of image and time series.

Image-based approaches typically utilize data from instruments like the Helioseismic and Magnetic Imager (HMI) and Atmospheric Imaging Assembly (AIA), focusing on the physical properties of active regions (E. Jonas et al. 2018; N. Nishizuka et al. 2021; P. Sun et al. 2022; Z. Sun et al. 2022; K. Yi et al. 2023; E. Amar & O. Ben-Shahar 2024; X. Zhang et al. 2024; D. Xu et al. 2025). Features such as shear angle, helicity, and magnetic energy are extracted from SHARP data (F. Ribeiro & A. Gradvohl 2021; Y. Abduallah et al. 2023; V. Deshmukh et al. 2023; P. A. Kosovich et al. 2024; X. Li et al. 2024; K. Saini et al. 2024; Y. Velanki et al. 2024; Y. Yang 2025), while advanced techniques like Zernike moments offer machine-interpretable descriptors from EUV images and magnetograms (A. Raboonik et al. 2016; N. Alipour et al. 2019). These methods aim to classify solar active regions as flaring or nonflaring, relying on spatial features like polarity inversion lines that signal imminent activity. However, the availability of high-resolution image data is limited to the post-SDO era (after 2010), restricting the historical depth of such studies.

In contrast, time series approaches utilize long-term records of solar X-ray flux observed by GOES. This data set spans several decades, enabling analysis over multiple solar cycles. However, challenges remain in distinguishing flare events from background variations, requiring specialized detection algorithms (B. Kaki et al. 2022).

Our study follows this approach, focusing on statistical and sequential patterns in flare activity rather than spatial characteristics. The method introduced in this work segments X-ray time series data using a sliding window technique to capture local temporal structures. DLSTM networks further enhance this analysis by decomposing time series into trend and seasonal components, reducing noise, and isolating flare-related patterns. When applied to regularized time series, DLSTM achieved a TSS of 0.68 and a recall of 0.90, which improved to 0.74 and 0.95, respectively, with an ensemble model. This demonstrates the strength of DLSTM in preserving quasi-periodic structures essential for predicting large flares. While LSTM-based models showed moderate performance (TSS = 0.41, recall = 0.51 on irregular data), resampling and regularization significantly enhanced their capability on structured time series (TSS = 0.68, recall = 0.89). However, oversampling beyond a threshold introduced overfitting, limiting further improvement.

A key challenge in solar flare forecasting is data set imbalance—large flares (M and X class) are rare compared to smaller events. This imbalance biases models toward frequent, less critical events. Using resampling techniques helps address this, though trade-offs like overfitting must be carefully managed.

Table 4 summarizes various machine learning methods used in flare prediction. Comparison across models is inherently difficult due to differences in data sets, time spans, data quality, and the physical parameters (G. Barnes et al. 2016; X. Huang et al. 2024). Image-based models and time series models not only differ in input types but also in learning objectives and underlying assumptions, complicating direct benchmarking.

Moreover, the implications of false predictions must not be overlooked. In space weather applications, FPs in large flare prediction or false negatives for small flares can have serious operational consequences, especially for satellite systems and military operations. Thus, improving model precision while minimizing false alarms remains a top priority.

Ultimately, both image-based and time series-based approaches offer unique strengths. While image-based models capture rich spatial features of active regions, time series models like ours provide long-term context and statistical depth. Integrating insights from both approaches may be the most promising path toward robust solar flare prediction.

## 6. Conclusion

This study explores applying advanced machine learning techniques, specifically LSTM and DLSTM models, for predicting solar flares. Notable contributions include the use of sliding window methods to identify temporal quasi-patterns, determining the optimal window size through statistical analysis, and employing resampling strategies to mitigate class imbalance in irregular and regular flare time series over 2003 January 11 to 2023 April 30 recorded by GOES. We constructed sliding windows for irregular (original) and regular time series to capture essential flare characteristics. For irregular time series, each sliding window included sequences of peak flux values and waiting times, while for regular time series, the windows contained sequences of maximum peak fluxes recorded at fixed 3 hr intervals, emphasizing the role of large flares and overall solar activity. During the flare time series regularization process, we excluded time gaps exceeding 24 hr from our analysis to distinguish active solar days from nonactive periods.

To evaluate forecasting performance, we implemented six different models: (1) LSTM applied to irregular time series, (2) DLSTM on irregular time series, (3) LSTM on regularized time series, (4) DLSTM on regularized time series, (5) LSTM on regularized time series incorporating ensemble learning, and (6) DLSTM on regularized time series with ensemble learning.

Figure 8 illustrates the flowchart for predicting small and large solar flares, which comprises three main steps: preprocessing (smoothing solar peak fluxes), applying a sliding window with resampling on both irregular and regularized flare time series, and leveraging a prediction approach. We applied LSTM and DLSTM models to the sliding windows of irregular time series. However, for the sliding windows of regularized flare time series, we utilized four models: LSTM, DLSTM, and their combinations with an ensemble approach.

Our findings consistently highlight the superior performance of the DLSTM model in capturing complex flare dynamics and enhancing classification accuracy. Key metrics such as TSS and AUC emphasize the role of resampling in improving model robustness, with the most significant gains observed at

**Table 4**
The Various Flare Prediction Works with Different Machine Learning Methods, Data Sets and Extracted Features, Period of Observations, and TSS Metric Values

| Article | Method | Data Set and Features | Period | Metrics (TSS) |
| --- | --- | --- | --- | --- |
| N. Alipour et al. (2019) | SVM classifier | AIA images and HMI magnetograms: invariant Zernike moments | 2010–2018 | 0.859 |
| F. Ribeiro & A. Gradvohl (2021) | SVM, random forest, and light gradient boosting machine | GOES and SHARP magnetograms: 37 X-ray flux, magnetic field parameters, and daily solar and particle parameters | 2001–2018 | 0.54 |
| Z. Sun et al. (2022) | CNN and an LSTM | HMI magnetograms | 2001–2018 | 0.54 |
| C. Pandey et al. (2022) | Logistic regression-based meta learner | Space weather analytics for solar flares data set | 2010–2018 | 0.52 |
| M. Li et al. (2022) | Pure CNN and two fusion models | SHARP magnetograms | 2010–2018 | 0.695 |
| Y. Abduallah et al. (2021) | Ensemble of random forest, multilayer perceptions, and extreme learning machines | SHARP magnetograms: 13 magnetic field parameters | 2010–2016 | 0.507 |
| P. Sun et al. (2022) | 3D CNNs | SHARP magnetograms: magnetic field components | 2010–2019 | 0.826 |
| V. Deshmukh et al. (2023) | Logistic regression, multilayer perceptrons, LSTMs, and extremely randomized trees | SHARP magnetograms: 20 magnetic field parameters | 2010–2017 | 0.79 |
| Y. Abduallah et al. (2023) | Transformer, CNN, and LSTM networks | SHARP magnetograms: nine magnetic field parameters | 2010–2022 | 0.835 |
| K. Yi et al. (2023) | Deep reinforcement learning | Heliospheric Observatory/Michelson Doppler Imager (1996–2010) and the HMI magnetograms (2011–2019) | 1996–2019 | 0.59 |
| E. Amar & O. Ben-Shahar (2024) | CNNs generating synthetic magnetograms to reduce the imbalance in the data | HMI magnetograms | 2011–2017 | 0.64 |
| S. Zhang et al. (2024) | Deep CNN | SHARP magnetograms | 2010–2018 | 0.764 |
| X. Zhang et al. (2024) | Causal attention deep learning model, AlexNet, SqueezeNet, and ResNet | HMI magnetograms | 2010–2019 | 0.7511 |
| K. Saini et al. (2024); Y. Velanki et al. (2024) | Minimally random convolutional kernel transform (MiniRocket), SVM, canonical interval forest, multiple representations sequence learner, LSTM, and Ensemble (random forest, MiniRocket) | Space weather analytics for solar flares data set: 24 parameters | 2010–2018 | ≈0.38, ≈0.58 |
| X. Li et al. (2024) | Transformer, BiLSTM-attention, LSTM, and NN models | SHARP magnetograms: 10 magnetic field parameters and high-energy-density magnetic field parameters (HED): six parameters of high-energy-density ARs | 2010–2022 | SHARP: 0.559, HED: 0.721 |
| D. Xu et al. (2025) | CNNs, a TCN, and CNN-TCN | HMI magnetograms: 16 parameters | 2010–2019 | 0.85 |
| This work | LSTM and DLSTM models with an ensemble learning | GOES flare catalog | 2003–2023 | 0.74 |

**Note.** For each study, we present the highest TSS value reported in the corresponding article.

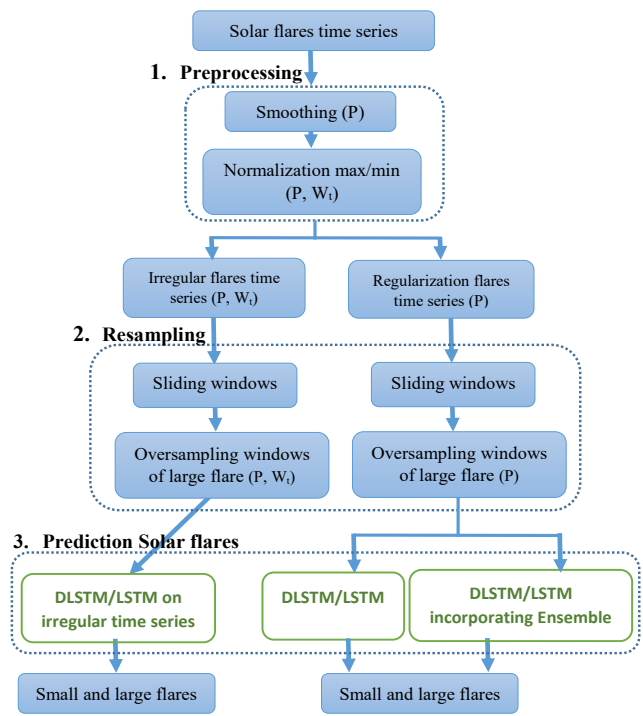

**Figure 8.** The flowchart for predicting small and large solar flares consists of three key steps: preprocessing (smoothing solar peak fluxes), applying a sliding window with resampling on both irregular and regular flare time series, and utilizing a prediction approach. During preprocessing, solar peak fluxes are smoothed, and peak fluxes ($P$) and waiting times ($W_t$) are normalized. Each sliding window comprises a peak flux subwindow and a waiting time subwindow, each containing a sequence of 24 flares. An oversampling algorithm is applied to address the imbalance in the minority class, corresponding to large (M and X) flares. For the prediction step, we employ six models: (1) LSTM and (2) DLSTM applied to irregular time series, (3) LSTM and (4) DLSTM applied to regular time series, and (5) LSTM and (6) DLSTM integrated with ensemble learning for regular time series.

moderate resampling levels. Despite the intricate nature of the solar flare system, the proportion of windows exhibiting "complex patterns" (unpredictable flare occurrences) remains relatively small. For instance, a total of 151,047 windows were generated, but only a minor fraction displayed irregular patterns, underscoring the inherent difficulty in forecasting solar flares over long periods.

As summarized in Table 3, DLSTM applied to regular time series with various resampling techniques emerged as the best-performing model, achieving the highest recall for large flares (0.95), an accuracy of approximately 80%, and the best overall skill scores. Ensemble learning improves forecasting performance, further solidifying DLSTM as the superior model. By capturing long-term dependencies and reducing noise through trend and seasonality extraction, DLSTM outperforms traditional LSTM, especially when combined with ensemble methods. This integration refines predictions and strengthens the model's ability to forecast large solar flares more accurately. Regularizing the time series with a 3 hr interval enhances model balance, leading to better classification results compared to irregular time series models. Resampling plays a crucial role in mitigating class imbalance, significantly boosting recall for large flares. Consequently, DLSTM on regular time series with resampling is the most effective approach for solar flare prediction.

These findings demonstrate the advantage of DLSTM in managing temporal dependencies and predicting flares despite the system's complexity. DLSTM outperforms LSTM primarily due to its ability to extract trends and seasonal components from time series data, allowing it to separate random noise more effectively than LSTM, which focuses on refining noise alone. Our results highlight the importance of integrating machine learning techniques with domain-specific knowledge to address the challenges of solar flare forecasting. The combination of DLSTM, sliding window techniques, resampling strategies, and robust evaluation metrics proves highly effective in classifying small and large flares.

Furthermore, the study emphasizes the need to incorporate various solar cycle phases and address class imbalance for more reliable forecasting. The challenge of predicting complex flare events is evident from the relatively few windows containing such patterns. Future research should explore integrating additional solar parameters, developing hybrid models, and extending predictive capabilities to improve preparedness for solar events and their potential impact on Earth's space weather environment.

### Acknowledgments

We thank the GOES science teams for providing the data used here. We also gratefully thank the anonymous referee for very helpful comments and suggestions that improved the manuscript.

### ORCID iDs

Zeinab Hassani https://orcid.org/0000-0001-5516-4377
Davud Mohammadpur https://orcid.org/0000-0002-1207-5830
Hossein Safari https://orcid.org/0000-0003-2326-3201

### References

Abda, Z. M. K., Aziz, N. F. A., Kadir, M. Z. A. A., & Rhazali, Z. A. 2020, IEEEA, 8, 200237
Abduallah, Y., Wang, J. T. L., Nie, Y., Liu, C., & Wang, H. 2021, RAA, 21, 160
Abduallah, Y., Wang, J. T. L., Wang, H., & Xu, Y. 2023, NatSR, 13, 13665
Ahmed, O. W., Qahwaji, R., Colak, T., et al. 2013, SoPh, 283, 157
Alipour, N., Mohammadi, F., & Safari, H. 2019, ApJS, 243, 20
Amar, E., & Ben-Shahar, O. 2024, ApJS, 271, 29
Asaly, S., Gottlieb, L.-A., & Reuveni, Y. 2021, IJSTA, 14, 1469
Aschwanden, M. 2011, Self-organized Criticality in Astrophysics: The Statistics of Nonlinear Processes in the Universe (Berlin: Springer)
Asensio Ramos, A., Cheung, M. C. M., Chifu, I., & Gafeira, R. 2023, LRSP, 20, 4
Baeke, H., Amaya, J., & Lapenta, G. 2023, ESS Open Archive, doi:10.22541/essoar.167336864.46114556/v1
Bar-Yam, Y. 1997, Dynamics of Complex Systems (Cambridge, MA: Perseus Books)
Barnes, G., Leka, K. D., Schrijver, C. J., et al. 2016, ApJ, 829, 89
Bishop, C. 2006, Pattern Recognition and Machine Learning (Berlin: Springer), 140
Bloomfield, D. S., Higgins, P. A., McAteer, R. T. J., & Gallagher, P. T. 2012, ApJL, 747, L41
Bobra, M. G., & Couvidat, S. 2015, ApJ, 798, 135
Boucheron, L. E., Al-Ghraibah, A., & McAteer, R. T. J. 2015, ApJ, 812, 51
Boucheron, L. E., Vincent, T., Grajeda, J. A., & Wuest, E. 2023, NatSD, 10, 825
Breiman, L. 1996, Mach. Learn., 24, 123
Chen, Q., Wen, D., Li, X., et al. 2019, PLoSO, 14, e0222365
Chen, W., Yang, K., Yu, Z., Shi, Y., & Chen, C. L. P. 2024, Artif. Intell. Rev., 57, 137
Chen, X., Wang, G., Zhou, W., Chang, S., & Sun, S. 2006, in Intelligent Computing, ed. D. S. Huang, K. Li, & G. W. Irwin (Berlin: Springer), 672
Chen, Y., Manchester, W. B., Hero, A. O., et al. 2019, SpWea, 17, 1404
Chou, J.-S., & Ngo, N.-T. 2016, ApEn, 177, 751

Colak, T., & Qahwaji, R. 2009, SpWea, 7, S06001
Crutchfield, J., & Young, K. 1990, in Complexity, Entropy and the Physics of Information, ed. W. H. Zurek (Reading, MA: Addison Wesley)
Dai, Q., Liu, J.-w., & Yang, J.-P. 2023, Eng. Appl. Artif. Intell., 121, 105959
Deshmukh, V., Baskar, S., Berger, T. E., Bradley, E., & Meiss, J. D. 2023, A&A, 674, A159
Dietterich, T. G. 2000, Multiple Classifier Systems (Berlin: Springer), 1
Dissauer, K., Leka, K. D., & Wagner, E. L. 2023, ApJ, 942, 83
Doswell, C. A., III, Davies-Jones, R., & Keller, D. L. 1990, WtFor, 5, 576
Eiglsperger, J., Haselbeck, F., & Grimm, D. G. 2023, Mach. Learn. Appl., 12, 100467
Estrada-Patiño, E., Castilla-Valdez, G., Frausto-Solis, J., González-Barbosa, J., & Sánchez-Hernández, J. P. 2024, Int. J. Comput. Intell. Syst., 17, 253
Farhang, N., Safari, H., & Wheatland, M. S. 2018, ApJ, 859, 41
Farhang, N., Shahbazi, F., & Safari, H. 2022, ApJ, 936, 87
Farhang, N., Wheatland, M. S., & Safari, H. 2019, ApJL, 883, L20
Florios, K., Kontogiannis, I., Park, S.-H., et al. 2018, SoPh, 293, 28
Foote, R. 2007, Sci, 318, 410
Georgoulis, M. K., Vilmer, N., & Crosby, N. B. 2001, A&A, 367, 326
Georgoulis, M. K., Yardley, S. L., Guerra, J. A., et al. 2024, AdSpR, in press
Gheibi, A., Safari, H., & Javaherian, M. 2017, ApJ, 847, 115
Grim, L. F. L., & Gradvohl, A. L. S. 2024, SoPh, 299, 33
Hochreiter, S., & Schmidhuber, J. 1997, Neural Comput., 9, 1735
Huang, X., Zhang, L., Wang, H., & Li, L. 2013, A&A, 549, 127
Huang, X., Zhao, Z., Zhong, Y., et al. 2024, ScChD, 67, 3727
Janka, D., Lenders, F., Wang, S., Cohen, A., & Li, N. 2019, Control Eng. Pract., 93, 104169
Jerse, G., & Marcucci, A. 2024, A&C, 46, 100786
Jonas, E., Bobra, M., Shankar, V., Todd Hoeksema, J., & Recht, B. 2018, SoPh, 293, 48
Kaki, B., Farhang, N., & Safari, H. 2022, NatSR, 12, 16835
Kontogiannis, I. 2023, AdSpR, 71, 2017
Kosovich, P. A., Kosovichev, A. G., Sadykov, V. M., et al. 2024, ApJ, 972, 169
Krista, L. D. 2025, ApJ, 980, 123
Krista, L. D., & Chih, M. 2021, ApJ, 922, 218
Landa, V., & Reuveni, Y. 2022, ApJS, 258, 12
LeCun, Y., Bengio, Y., & Hinton, G. 2015, Natur, 521, 436
Leka, K. D., Dissauer, K., Barnes, G., & Wagner, E. L. 2023, ApJ, 942, 84
Li, L., & Spratling, M. 2023, PatRe, 136, 109229
Li, M., Cui, Y., Luo, B., et al. 2022, SpWea, 20, e2021SW002985
Li, X., Li, X., Zheng, Y., et al. 2024, ApJS, 276, 7
Liu, C., Deng, N., Wang, J. T. L., & Wang, H. 2017, ApJ, 843, 104
Liu, J., Yao, J., Zhou, Q., Wang, Z., & Huang, L. 2023, Appl. Intell., 53, 21077
Liu, S., Wang, J., Li, M., et al. 2023, FrASS, 10, 1082694
Liu, X., Du, H., & Yu, J. 2023, Neurocomputing, 556, 126648
Lotfi, N., Javaherian, M., Kaki, B., Darooneh, A. H., & Safari, H. 2020, Chaos, 30, 043124
MacKay, R. S. 2008, Nonli, 21, T273
McIntosh, P. S. 1990, SoPh, 125, 251
Mngomezulu, M., Gwetu, M., & Fonou-Dombeu, J. V. 2023, in Artificial Intelligence XL, ed. M. Bramer & F. Stahl (Cham: Springer), 307
Mohammed, A., & Kora, R. 2023, J. King Saud Univ.–Comput. Inform. Sci., 35, 757
Moniz, N., Branco, P., & Torgo, L. 2017, Int. J. Data Sci. Anal., 3, 161
Nagem, T., Qahwaji, R., Ipson, S., Wang, Z., & Al-Waisy, A. 2018, Int. J. Adv. Comput. Sci. Appl., 9, 492
Nishizuka, N., Kubo, Y., Sugiura, K., Den, M., & Ishii, M. 2020, ApJ, 899, 150
Nishizuka, N., Kubo, Y., Sugiura, K., Den, M., & Ishii, M. 2021, EP&S, 73, 64
Nishizuka, N., Sugiura, K., Kubo, Y., et al. 2017, ApJ, 835, 156
Oliveira, D. M., & Ngwira, C. M. 2017, BrJPh, 47, 552
Ozawa, K., Itakura, T., & Ono, T. 2024, ApSpe, 78, 825
Pandey, C., Ji, A., Angryk, R. A., Georgoulis, M. K., & Aydin, B. 2022, FrASS, 9, 2022
Plutino, N., Berrilli, F., Del Moro, D., & Giovannelli, L. 2023, AdSpR, 71, 2048
Raboonik, A., Safari, H., Alipour, N., & Wheatland, M. S. 2016, ApJ, 834, 11
Ribeiro, F., & Gradvohl, A. 2021, A&C, 35, 100468
Saini, K., Alshammari, K., Hamdi, S. M., & Filali Boubrahimi, S. 2024, Univ, 10, 234
Salman, D., Direkoglu, C., Kusaf, M., & Fahrioglu, M. 2024, Neural Comput. Appl., 36, 9095
Schapire, R. E. 1990, Mach. Learn., 5, 197
Sen, R., Yu, H.-F., & Dhillon, I. S. 2019, Advances in Neural Information Processing Systems 32, ed. H. Wallach et al. (NeurIPS), https://proceedings.neurips.cc/paper_files/paper/2019/file/3a0844cee4fcf57de0c71e9ad3035478-Paper.pdf
Shan, F., He, X., Armaghani, D. J., & Sheng, D. 2024, JRMGE, 16, 1538
Shin, S., Lee, J.-Y., Moon, Y.-J., Chu, H., & Park, J. 2016, SoPh, 291, 897
Sinha, S., Gupta, O., Singh, V., et al. 2022, ApJ, 935, 45
Song, H., Tan, C., Jing, J., et al. 2009, SoPh, 254, 101
Strugarek, A., & Charbonneau, P. 2014, SoPh, 289, 4137
Sun, P., Dai, W., Ding, W., et al. 2022, ApJ, 941, 1
Sun, Z., Bobra, M. G., Wang, X., et al. 2022, ApJ, 931, 163
Tajik, Z., Farhang, N., Safari, H., & Wheatland, M. S. 2024, ApJS, 273, 1
Tang, R., Liao, W., Chen, Z., et al. 2021, ApJS, 257, 50
Taran, S., Alipour, N., Rokni, K., et al. 2023, AdSpR, 71, 5453
Taran, S., Khodakarami, E., & Safari, H. 2022, AdSpR, 70, 2541
Taylor, S. J., & Letham, B. 2018, Am. Stat., 72, 37
Thibeault, C., Strugarek, A., Charbonneau, P., & Tremblay, B. 2022, SoPh, 297, 125
Vafaeipour, M., Rahbari, O., Rosen, M. A., Fazelpour, F., & Ansarirad, P. 2014, IJEEE, 5, 105
Velanki, Y., Hosseinzadeh, P., Boubrahimi, S. F., & Hamdi, S. M. 2024, Univ, 10, 373
Vysakh, P. A., & Mayank, P. 2023, SoPh, 298, 137
Wang, C., Li, Y., Sun, X., et al. 2023, Proc. Thirty-Second Int. Joint Conf. on Artificial Intelligence, IJCAI-23, ed. E. Elkind, (International Joint Conferences on Artificial Intelligence Organization), 4299
Wang, G., Li, Q., Wang, L., et al. 2018, Sensors, 18, 1695
Wang, J., Luo, B., Liu, S., & Zhang, Y. 2023, ApJS, 269, 54
Wang, J., Zhang, Y., Webber, S. A. H., et al. 2020, ApJ, 892, 140
Wang, M., Zhang, L., Yu, H., et al. 2023, Comput. Biol. Med., 159, 106879
Wheatland, M. S. 2005, SpWea, 3, S07003
Wheatland, M. S., & Litvinenko, Y. E. 2002, SoPh, 211, 255
Wibawa, A. P., Utama, A. B. P., Elmunsyah, H., et al. 2022, J. Big Data, 9, 44
Wu, B., Fang, C., Yao, Z., Tu, Y., & Chen, Y. 2023, Electronics, 12, 354
Wu, H., Xu, J., Wang, J., & Long, M. 2021, Advances in Neural Information Processing Systems 34, ed. M. Ranzato et al. (NeurIPS), 22419, https://proceedings.neurips.cc/paper/2021/hash/bcc0d400288793e8bdcd7c19a8ac0c2b-Abstract.html
Xu, D., Sun, P., Feng, S., Liang, B., & Dai, W. 2025, ApJS, 276, 68
Yang, Y. 2025, FrASS, 11, 1509061
Yi, K., Moon, Y.-J., & Jeong, H.-J. 2023, ApJS, 265, 34
Yu, D., Huang, X., Wang, H., & Cui, Y. 2009, SoPh, 255, 91
Yuan, Y., Shih, F. Y., Jing, J., & Wang, H.-M. 2010, RAA, 10, 785
Zhang, G., Wang, P., & Geng, X. 2024, in Proc. 2024 Int. Academic Conf. on Edge Computing, Parallel and Distributed Computing, ECPDC '24 (New York: Association for Computing Machinery), 226
Zhang, K., Ning, W., Zhu, Y., et al. 2024, ApSci, 14, 218
Zhang, S., Zheng, Y., Li, X., et al. 2024, AdSpR, 74, 3467
Zhang, X., Xu, L., Li, Z., & Huang, X. 2024, ApJS, 274, 38